\documentclass[a4paper]{article}

\usepackage{INTERSPEECH2021}
\usepackage[normalem]{ulem}
\usepackage[dvipsnames]{xcolor}
\usepackage[T1]{fontenc}
\usepackage{bm}
\useunder{\uline}{\ul}{}
\PassOptionsToPackage{hyphens}{url}\usepackage{hyperref}
\title{Multi-speaker Emotional Text-to-speech Synthesizer}
\name{
    Sungjae Cho$^{1,\dagger}$, 
    Soo-Young Lee$^2$
    \thanks{$^{\dagger}$ work done at KAIST}
}

\address{
    $^1$Korea Institute of Science and Technology, Republic of Korea\\
    $^2$Korea Advanced Institute of Science and Technology, Republic of Korea
}
\email{
    sj.cho@snu.ac.kr,
    sylee@kaist.ac.kr
}

\begin{document}

\maketitle

\begin{abstract}
We present a methodology to train our multi-speaker emotional text-to-speech synthesizer that can express speech for 10 speakers' 7 different emotions.
All silences from audio samples are removed prior to learning.
This results in fast learning by our model.
Curriculum learning is applied to train our model efficiently.
Our model is first trained with a large single-speaker neutral dataset, and then trained with neutral speech from all speakers.
Finally, our model is trained using datasets of emotional speech from all speakers.
In each stage, training samples of each speaker-emotion pair have equal probability to appear in mini-batches.
Through this procedure, our model can synthesize speech for all targeted speakers and emotions.
Our synthesized audio sets are available on our web page.
\end{abstract}

\section{Introduction}

Emotional speech synthesis has been achieved through deep neural networks~\cite{emoTTS_LeeRL17, emoTTS_ChoiPPH19, emoTTS_UmOBJAK20,emoTTS_KimCCPL20}.
However, most studies have trained models on a small number of speakers or balanced class distributions because it is challenging to guarantee speech quality for each speaker and emotion, given imbalanced data distributions with respect to speakers and emotions.
In this paper, we present a methodology for training our multi-speaker emotional text-to-speech (TTS) synthesizer capable of generating speech for all targeted speakers' voices and emotions. 
The main methods are silence removal, curriculum learning~\cite{curriculum-learning_BengioLCW09}, and oversampling~\cite{oversampling_BudaMM18}. 
The synthesized audios are demonstrated through a web page.

\section{Datasets}

4 datasets were used to train the multi-speaker emotional TTS synthesizer.
The first dataset, the Korean single speaker speech (KSS) dataset~\cite{KSS}, is publicly available and contains speech samples of a single female speaker: kss-f.
We labeled their emotion as neutral.
The remaining 3 datasets consist of speech of 
the Ekman's 7 basic emotions~\cite{7emotions_ekman2011}: neutral, anger, disgust, fear, happiness, sadness, and surprise.

The first Korean emotional TTS (KETTS) dataset consists of 1 female and 1 male speaker: ketts-30f and ketts-30m, which are abbreviations of a 30's female and male in KETTS.
The 2 speakers were assigned to different sets of sentences; however, the same sentences were recorded across 7 emotions for a single speaker.
In the female case, only happy speech samples have a different set of sentences.
KETTS is balanced with respect to speakers and emotions, except for the female's happy speech subset (Table \ref{table/datasets}).

The second Korean emotional TTS (KETTS2) dataset consists of 3 female and 3 male speakers, totally 6 speakers: kett2-20m, ketts2-30f, ketts2-40m, ketts-50f, ketts2-50m, and ketts2-60f.
The same sentences were recorded across 7 emotions and 6 speakers.
Hence, KETTS2 is balanced with respect to speakers and emotions (Table \ref{table/datasets}).

The third Korean emotional TTS (KETTS3) dataset consists of 1 female and 1 male speaker: ketts3-f and ketts3-m.
It includes 5 emotions, excluding disgust and surprise.
The same sentences were recorded across 2 speakers; however, different sentences were spoken for the 5 emotions.
KETTS3 is balanced for speakers but not for emotions.
Therefore, the whole training dataset is balanced for neither speakers nor emotions (Table \ref{table/datasets}).

\begin{table}[!t]
\scriptsize
\setlength\tabcolsep{3.5pt}
\centering
\begin{center} 
\caption{
    Hours of preprocessed training datasets
}
\label{table/datasets} 
\begin{tabular}{lrrrrrrrr}
\toprule
\multicolumn{1}{c}{\textbf{Speaker}} & \multicolumn{1}{c}{\textbf{all}} & \multicolumn{1}{c}{\textbf{neu}} & \multicolumn{1}{c}{\textbf{ang}} & \multicolumn{1}{c}{\textbf{dis}} & \multicolumn{1}{c}{\textbf{fea}} & \multicolumn{1}{c}{\textbf{hap}} & \multicolumn{1}{c}{\textbf{sad}} & \multicolumn{1}{c}{\textbf{sur}} \\

\midrule
kss-f                       & 12.59                   & 12.59                   &                         &                         &                         &                         &                         &                         \\
\midrule
ketts-30f                   & 26.61                   & 3.52                    & 3.46                    & 3.51                    & 3.68                    & 5.13                    & 3.75                    & 3.56                    \\
ketts-30m                   & 24.12                   & 3.37                    & 3.29                    & 3.31                    & 3.51                    & 3.50                    & 3.73                    & 3.40                    \\
\midrule
ketts2-20m                  & 5.09                    & 0.72                    & 0.72                    & 0.74                    & 0.76                    & 0.69                    & 0.75                    & 0.70                    \\
ketts2-30f                  & 4.69                    & 0.66                    & 0.65                    & 0.67                    & 0.65                    & 0.70                    & 0.68                    & 0.68                    \\
ketts2-40m                  & 4.98                    & 0.73                    & 0.69                    & 0.70                    & 0.75                    & 0.69                    & 0.74                    & 0.69                    \\
ketts2-50f                  & 4.98                    & 0.73                    & 0.71                    & 0.71                    & 0.70                    & 0.72                    & 0.71                    & 0.69                    \\
ketts2-50m                  & 4.73                    & 0.68                    & 0.68                    & 0.69                    & 0.67                    & 0.68                    & 0.68                    & 0.65                    \\
ketts2-60f                  & 4.90                    & 0.77                    & 0.68                    & 0.67                    & 0.68                    & 0.72                    & 0.72                    & 0.67                    \\
\midrule
ketts3-f                    & 9.64                    & 3.96                    & 1.34                    &                         & 1.27                    & 1.44                    & 1.64                    &                         \\
ketts3-m                    & 9.38                    & 3.90                    & 1.43                    &                         & 1.18                    & 1.39                    & 1.48                    &                         \\
\midrule
all                         & 111.70                  & 31.63                   & 13.65                   & 11.01                   & 13.85                   & 15.64                   & 14.87                   & 11.05                    \\
\bottomrule
\end{tabular}
\end{center}
\end{table}

\section{Methodology}

\subsection{Preprocessing}

The WebRTC voice activity detector, py-webrtcvad\footnote{\url{https://github.com/wiseman/py-webrtcvad}}, is utilized to remove unvoiced segments in audios, with its settings of an aggressiveness level of 3, frame duration 30ms, and padding duration 150ms.
These settings remove silences at the start, end, and middle of speech.
However, the amount of silence removed does not distort emotional expression.
All audios are resampled to sampling rate 22,050Hz.
Mel spectrograms are computed through a short-time Fourier transform (STFT) using frame size 1024, hop size 256, window size 1024, and a Hann window function.
The STFT magnitudes are transformed to the librosa Slaney mel scale using an 80-channel mel filterbank spanning 0Hz to 8kHz, and the results are then clipped to a minimum value of $10^{-5}$, followed by log dynamic range compression.

Every Korean character in an input sentence is decomposed into 3 elements: an onset, nucleus, and coda.
In total, 19 onsets, 21 nuclei, and 28 codas including the empty coda are employed as defined by Unicode.
A sequence of these elements becomes a grapheme sequence taken as input by our synthesizer.

\subsection{Model}

Our multi-speaker emotional TTS synthesizer takes 3 inputs --- the grapheme sequence of a Korean sentence, 1 of 10 speakers (5 females, 5 males), and 1 of the 7 Ekman's emotion classes.
It then generates a waveform in which the speaker utters the input sentence with the given emotion.
Our synthesizer consists of 2 sub-models: \textit{Tacotron 2}~\cite{tacotron2_ShenPWSJYCZWRSA18}, mapping a grapheme sequence to a mel spectrogram, and \textit{WaveGlow}~\cite{waveglow_PrengerVC19}, transforming the mel spectrogram to a waveform.
Tacotron 2 is an auto-regressive sequence-to-sequence neural network with a location-sensitive attention mechanism.
WaveGlow is a flow-based generative neural network without auto-regression.
We adapted NVIDIA Tacotron 2 and WaveGlow repositories\footnote{\url{https://github.com/NVIDIA/tacotron2}}$^,$\footnote{\url{https://github.com/NVIDIA/waveglow}} to synthesize speech for multiple speakers and emotions.
The WaveGlow model was utilized without modification but the Tacotron 2 model was modified as outlined in the following paragraph.

Speaker identity is represented as a 5-dimensional trainable \textit{speaker vector}.
Emotion identity is represented as a 3-dimensional trainable \textit{emotion vector}, except for the neutral emotion vector, which is a non-trainable zero vector.
To synthesize speech of a given speaker and emotion, in the decoder of Tacotron 2, speaker and emotion vectors are concatenated to attention context vectors taken by the first and second LSTM layers and the linear layer estimating a mel spectrogram.

\subsection{Training}

Tacotron 2 was trained with a batch size of 64 equally distributed to 4 GPUs.
The Adam optimizer~\cite{adam_KingmaB14} of the default settings ($\beta_1=0.9$, $\beta_2=0.999$, $\epsilon=10^{-6}$) was used with a learning rate of $10^{-3}$ and $L_2$ regularization with weight $10^{-6}$.
If the norm of gradients exceeded 1, their norm was normalized to 1 to ensure stable learning.

Using a curriculum learning~\cite{curriculum-learning_BengioLCW09} strategy, Tacotron 2 was trained to learn
single-speaker neutral speech, multi-speaker neutral speech, and multi-speaker emotional speech in this order.
More specifically, the model was trained with the KSS dataset for 20,000 iterations, then additionally with all datasets of neutral speech for 30,000 iterations, and finally with all training datasets for 65,000 iterations.
Transitioning to training on the next dataset was done when the model stably pronounced given whole sentences for all training speaker-emotion pairs.

In each training stage, we oversampled~\cite{oversampling_BudaMM18} the training set with respect to speaker-emotion pairs, which means samples of each speaker-emotion pair appear in a mini-batch with equal probability.
For example, samples of (ketts-30f, neutral) and those of (ketts2-20m, happy) appear in a mini-batch with equal probability.
This helped overcome difficulty in learning to synthesize speech of speaker-emotion pairs with relatively scarce samples.

WaveGlow was trained with a batch size of 24, equally distributed to 3 GPUs using 24 clips of 16,000 mel spectrogram frames randomly chosen from each training sample.
Training samples shorter than 16,000 mel frames were excluded from the training set since these samples padded with zeros caused unstable learning such as exploding gradients.
Similar to Tacotron 2, we oversampled the training set with respect to speaker-emotion pairs.
The Adam optimizer was used with the default settings and learning rate $10^{-4}$.
Weight normalization was applied, as described in the original paper~\cite{waveglow_PrengerVC19}.
To ensure stable learning, if the norm of gradients exceeded 1, their norm was normalized to 1.
The model was initialized with the pretrained weights\footnote{``waveglow\_256channels\_universal\_v5.pt'' was used.} offered in the WaveGlow repository.
The network was trained for 400,000 iterations until its loss curve formed a plateau.
The $\bm{z}$ elements were sampled from Gaussians with standard deviation 1 during training and 0.75 during inference.
    
\section{Results and Discussion}

Through this procedure, our speech synthesizer is able to synthesize speech for all available 10 speakers and 7 emotions. 
Unexpectedly, disgusted and surprised expressions of the KETTS3 speakers can be synthesized even without training supervision.
Synthesized speech samples can be found on
this web page\footnote{\url{https://sungjae-cho.github.io/InterSpeech2021\_STDemo/}}.

Although our model expresses speaker and emotion identities,
there are some minor inconsistencies in the quality of synthesized samples across speakers and emotions.
Thus, in production, it is reasonable to fine-tune for each speaker and respectively preserve the model parameters.

Our silence removal settings substantially accelerated the learning of Tacotron 2.
This was probably because silence removal at the start, end, and middle of speech resulted in the linear relationship between text and speech, and this relationship helped the location-sensitive attention network easily learn text-to-speech alignments.

\section{Acknowledgements}

This work was supported by \textit{Ministry of Culture, Sports and Tourism} and \textit{Korea Creative Content Agency} [R2019020013, R2020040298].

\bibliographystyle{IEEEtran}

\bibliography{main}

\end{document}